\documentclass[10pt,twocolumn,letterpaper]{article}

\usepackage{iccv}
\usepackage{times}
\usepackage{epsfig}
\usepackage{graphicx}
\usepackage{amsmath}
\usepackage{amssymb}

\usepackage[breaklinks=true,bookmarks=false]{hyperref}

\iccvfinalcopy 

\def\iccvPaperID{294} 

\ificcvfinal\fi

\begin{document}

\title{CAMP: Cross-Modal Adaptive Message Passing for Text-Image Retrieval}

\author{Zihao Wang$^{1*}$~~
Xihui Liu$^{1}$\thanks{The first two authors contributed equally to this work.}~~
Hongsheng Li$^1$~~
Lu Sheng$^3$~~
Junjie Yan$^2$~~
Xiaogang Wang$^1$~~
Jing Shao$^2$
\\
$^1$CUHK-SenseTime Joint Lab, The Chinese University of Hong Kong\\
$^2$SenseTime Research~~~~$^3$Beihang University\\
{\tt\small zihaowang@cuhk.edu.hk}~~~~~~
{\tt\small\{xihuiliu, hsli, xgwang\}@ee.cuhk.edu.hk}\\
{\tt\small lsheng@buaa.edu.cn}~~~~~~{\tt\small \{yanjunjie, shaojing\}@sensetime.com}
}

\maketitle
\ificcvfinal\thispagestyle{empty}\fi


\begin{abstract}
Text-image cross-modal retrieval is a challenging task in the field of language and vision. Most previous approaches independently embed images and sentences into a joint embedding space and compare their similarities. However, previous approaches rarely explore the interactions between images and sentences before calculating similarities in the joint space. Intuitively, when matching between images and sentences, human beings would alternatively attend to regions in images and words in sentences, and select the most salient information considering the interaction between both modalities. In this paper, we propose Cross-modal Adaptive Message Passing (CAMP), which adaptively controls the information flow for message passing across modalities. Our approach not only takes comprehensive and fine-grained cross-modal interactions into account, but also properly handles negative pairs and irrelevant information with an adaptive gating scheme. Moreover, instead of conventional joint embedding approaches for text-image matching, we infer the matching score based on the fused features, and propose a hardest negative binary cross-entropy loss for training. Results on COCO and Flickr30k significantly surpass state-of-the-art methods, demonstrating the effectiveness of our approach. \footnote{\url{https://github.com/ZihaoWang-CV/CAMP_iccv19}} 
\end{abstract}

\vspace{-10pt}
\section{Introduction}

Text-image cross-modal retrieval has made great progress recently~\cite{lee2018stacked,gu2018look,nam2016dual,faghri2017vse++,eisenschtat2017linking}. 
Nevertheless, matching images and sentences is still far from being solved, because of the large visual-semantic discrepancy between language and vision. 
Most previous work exploits visual-semantic embedding, which independently embeds images and sentences into the same embedding space, and then measures their similarities by feature distances in the joint space~\cite{karpathy2015deep,faghri2017vse++}. 
The model is trained with ranking loss, which forces the similarity of positive pairs to be higher than that of negative pairs. 
However, such independent embedding approaches do not exploit the interaction between images and sentences, which might lead to suboptimal features for text-image matching.

\begin{figure}[t]
\begin{center}
   \includegraphics[width=1\linewidth]{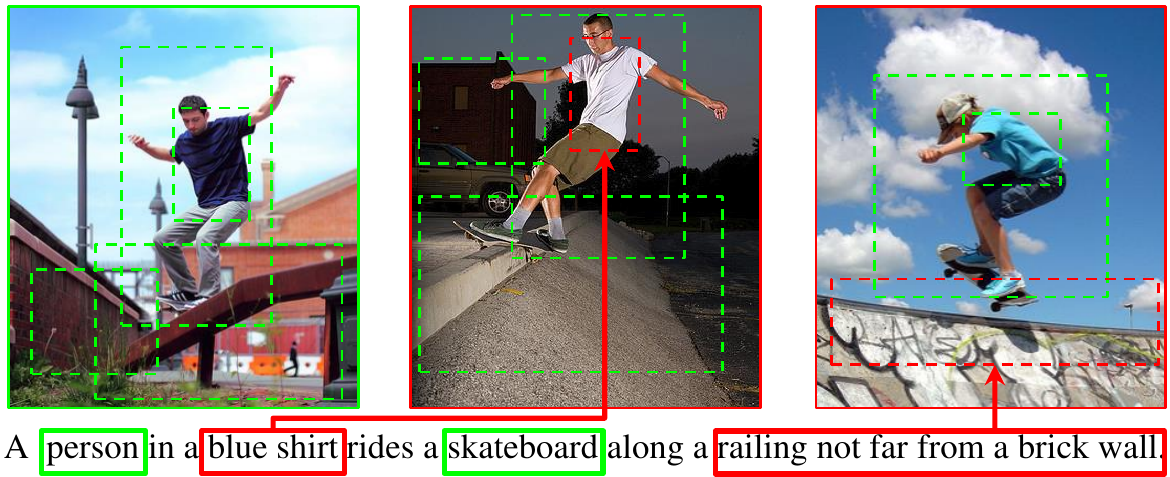}
\end{center}
\vspace{-10pt}
   \caption{Illustration of how our model distinguish the subtle differences by cross-modal interactions. Green denotes positive evidence, while red denotes negative cross-modal evidence.}
\label{fig:intro_compare}
\vspace{-17pt}
\end{figure}

Let us consider how we would perform the task of text-image matching ourselves. 
Not only do we concentrate on salient regions in the image and salient words in the sentence, but also we would alternatively attend to information from both modalities, take the interactions between regions and words into consideration, filter out irrelevant information, and find the fine-grained cues for cross-modal matching.
For example, in Figure~\ref{fig:intro_compare}, all of the three images seem to match with the sentence at first glance. 
When we take a closer observation, however, we would notice that the sentence describes ``blue shirt'' which cannot be found in the second image. 
Similarly, the description of ``a railing not far from a brick wall'' cannot be found in the third image. 
Those fine-grained misalignments can only be noticed if we have a gist of the sentence in mind when looking at the images.
As a result, incorporating the interaction between images and sentences should benefit in capturing the fine-grained cross-modal cues for text-image matching. 

%
%
In order to enable interactions between images and sentences, we introduce a \textit{\textbf{C}ross-modal \textbf{A}daptive \textbf{M}essage \textbf{P}assing} model (CAMP), composed of the \textit{Cross-modal Message Aggregation} module and the \textit{Cross-modal Gated Fusion} module. 
Message passing for text-image retrieval is non-trivial and essentially different from previous message passing approaches, mainly because of the existing of negative pairs for matching. 
If we pass cross-modal messages between negative pairs and positive pairs in the same manner, the model would get confused and it would be difficult to find alignments that are necessary for matching.
Even for matched images and sentences, information unrelated to text-image matching (\eg, background regions that are not described in the sentence) should also be suppressed during message passing.
Hence we need to adaptively control to what extent the messages from the other modality should be fused with the original features.
We solve this problem by exploiting a soft gate for fusion to adaptively control the information flow for message passing.
%

The \textbf{Cross-Modal Message Aggregation module} aggregates salient visual information corresponding to each word as messages passing from visual to textual modality, and aggregates salient textual information corresponding to each region as messages from textual to visual modality. 
The Cross-modal Message Aggregation is done by cross-modal attention between words and image regions. 
Specifically, we use region features as cues to attend on words, and use word features as cues to attend on image regions.
In this way, we interactively process the information from visual and textual modalities in the context of the other modality, and aggregate salient features as messages to be passed across modalities.
Such a mechanism considers the word-region correspondences and empowers the model to explore the fine-grained cross-modal interactions.
After aggregating messages from both modalities, the next step is fusing the original features with the aggregated messages passed from the other modality.
Despite the success of feature fusion in other problems such as visual question answering~\cite{fukui2016multimodal,gao2016compact,kim2016hadamard,yu2017multi,nguyen2018improved}, cross-modal feature fusion for text-image retrieval is nontrivial and has not been investigated before. 
In visual question answering, we only fuse the features of images and corresponding questions which are matched to the images.
For text-image retrieval, however, the key challenge is that the input image-sentence pair does not necessarily match. 
If we fuse the negative (mismatched) pairs, the model would get confused and have trouble figuring out the misalignments.
Our experiments indicate that na\"ive fusion approach does not work for text-image retrieval. 
To filter out the effects of negative (mismatched) pairs during fusion, we propose a novel \textbf{Cross-modal Gated Fusion module} to adaptively control the fusion intensity. 
Specifically, when we fuse the original features from one modality with the aggregated message passed from another modality, a soft gate adaptively controls to what extent the information should be fused. 
The aligned features are fused to a larger extent. 
While non-corresponding features are not intensively fused, and the model would preserve original features for negative pairs. 
The Cross-modal Gated Fusion module incorporates deeper and more comprehensive interactions between images and sentences, and appropriately handles the effect of negative pairs and irrelevant background information by an adaptive gate.

With the fused features, a subsequent question is: how to exploit the fused cross-modal information to infer the text-image correspondences? 
Since we have a joint representation consisting of information from both images and sentences, the assumption that visual and textual features are respectively embedded into the same embedding space no longer holds. 
As a result, we can no longer calculate the feature distance in the embedding space and train with ranking loss. 
%
%
%
We directly predict the cross-modal matching score based on the fused features, and exploit binary cross-entropy loss with hardest negative pairs as training supervision. 
%
Such reformulation gives better results, and we believe that it is superior to embedding cross-modal features into a joint space. 
%
%
By assuming that features from different modalities are separately embedded into the joint space, visual semantic embedding naturally prevents the model from exploring cross-modal fusion.
On the contrary, our approach is able to preserve more comprehensive information from both modalities, as well as fully exploring the fine-grained cross-modal interactions. 

To summarize, we introduce a Cross-modal Adaptive Message Passing model, composed of the Cross-modal Message Aggregation module and the Cross-modal Gated Fusion module, to adaptively explore the interactions between images and sentences for text-image matching. Furthermore, we infer the text-image matching score based on the fused features, and train the model by a hardest negative binary cross-entropy loss, which provides an alternative to conventional visual-semantic embedding. Experiments on COCO~\cite{lin2014microsoft} and Flickr30k~\cite{karpathy2015deep} validate the effectiveness of our approach.




\begin{figure*}[t]
\begin{center}
   \includegraphics[width=1\linewidth]{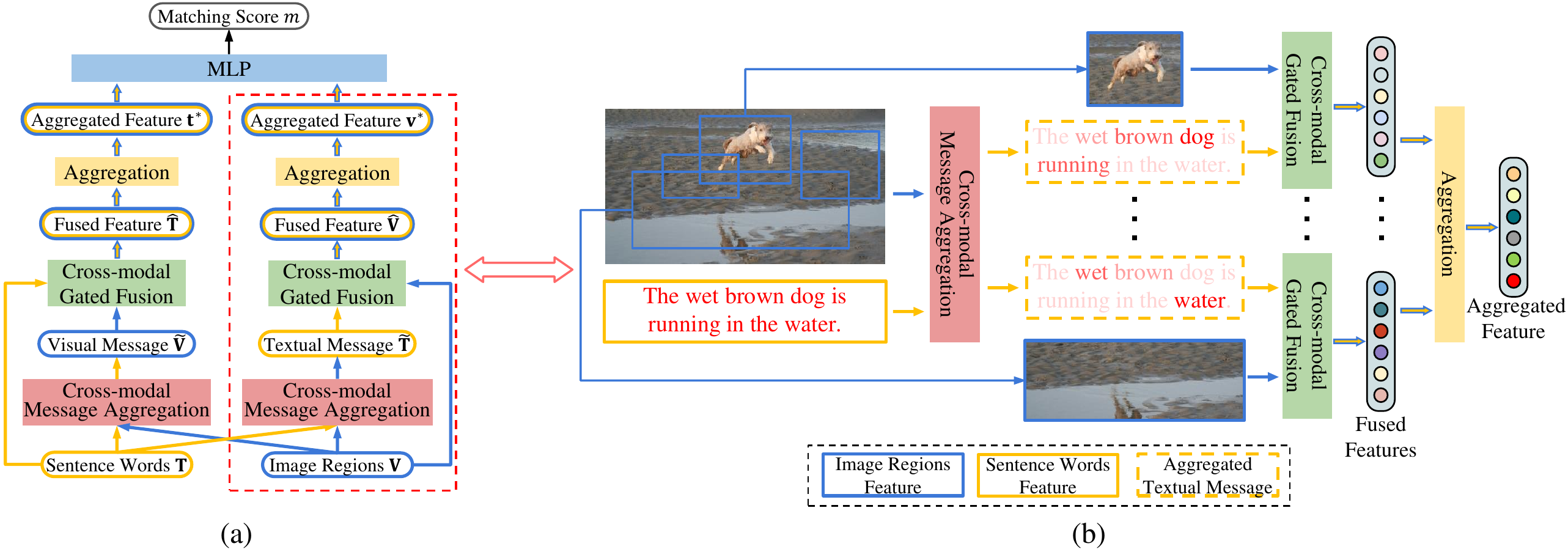}
\end{center}
\vspace{-10pt}
   \caption{(a) is the overview of the Cross-modal Adaptive Message Passing model. The input regions and words interact with each other and are aggregated to fused features to predict the matching score. (b) is an illustration of the message passing from textual to visual modality (the dashed red box in (a)). Word features are aggregated based on the cross-modal attention weights, and the aggregated textual messages are passed to fuse with the region features. The message passing from visual to textual modality operates in a similar way.}
\label{fig:three_models}
\vspace{-10pt}
\end{figure*}

\section{Related Work}

\noindent\textbf{Text-image retrieval.}
Matching between images and sentences is the key to text-image cross-modal retrieval. 
Most previous works exploited visual-semantic embedding to calculate the similarities between image and sentence features after embedding them into the joint embedding space, which was usually trained by ranking loss~\cite{kiros2014unifying,vendrov2015order,wang2016learning,klein2015associating,fang2015captions,eisenschtat2017linking,plummer2017conditional,karpathy2015deep}. 
Faghri~\etal~\cite{faghri2017vse++} improved the ranking loss by introducing the hardest negative pairs for calculating loss. 
Zheng~\etal~\cite{zheng2017dual} explored text CNN and instance loss to learn more discriminative embeddings of images and sentences. 
Zhang~\etal~\cite{ying2018CMPM} used projection classification loss which categorized the vector projection of representations from one modality onto another with the improved norm-softmax loss.
Niu~\etal~\cite{niu2017hierarchical} exploited a hierarchical LSTM model for learning visual-semantic embedding. 
Huang~\etal~\cite{huang2017learning} proposed a model to learn semantic concepts and order for better image and sentence matching. 
Gu~\etal~\cite{gu2018look} leveraged generative models to learn concrete grounded representations that capture the detailed similarity between the two modalities. 
%
%
Lee~\etal~\cite{lee2018stacked} proposed stacked cross attention to exploit the correspondences between words and regions for discovering full latent alignments. 
Nevertheless, the model only attends to either words or regions, and it cannot attend to both modalities symmetrically.
Different from previous methods, our model exploits cross-modal interactions by adaptive message passing to extract the most salient features for text-image matching.

\noindent\textbf{Interactions between language and vision.}
Different types of interactions have been explored in language and vision tasks beyond text-image retrieval~\cite{yu2017multi,chen2017sca,lu2017knowing,zhu2017structured,kim2018bilinear,xu2016ask,lu2016hierarchical,liu2018show,liu2019improving}. 
Yang~\etal~\cite{yang2016stacked} proposed stacked attention networks to perform multiple steps of attention on image feature maps. 
Anderson~\etal~\cite{anderson2017bottom} proposed bottom-up and top-down attention to attend to uniform grids and object proposals for image captioning and visual question answering (VQA). 
%
%
Previous works also explored fusion between images and questions~\cite{fukui2016multimodal,gao2016compact,kim2016hadamard,yu2017multi,nguyen2018improved} in VQA. 
%
%
Despite the great success in other language and vision tasks, few works explore the interactions between sentences and images for text-image retrieval, where the main challenge is to properly handle the negative pairs.
%
%
To our best knowledge, this is the first work to explore deep cross-modal interactions between images and sentences for text-image retrieval.

\vspace{-5pt}
\section{Cross-modal Adaptive Message Passing}
\label{sec:interaction_strategy}



In this section, we introduce our Cross-modal Adaptive Message Passing model to enable deep interactions between images and sentences, as shown in Fig.~\ref{fig:three_models}. 
The model is composed of two modules, \textit{Cross-modal Message Aggregation} and \textit{Cross-modal Gated Fusion}. 
Firstly we introduce the Cross-modal Message Aggregation based on cross-modal attention, and then we consider fusing the original information with aggregated messages passed from the other modality, which is non-trivial because fusing the negative (mismatched) pairs makes it difficult to find informative alignments. 
We introduce our Cross-modal Gated Fusion module to adaptively control the fusion of aligned and misaligned information.

\vspace{0pt}
\noindent\textbf{Problem formulation and notations.} 
Given an input sentence $\mathcal{C}$ and an input image $\mathcal{I}$, we extract the word-level textual features $\mathbf{T} = [\mathbf{t}_1, \cdots, \mathbf{t}_N] \in \mathbb{R}^{d \times N}$ for $N$ words in the sentence and region-level visual features $\mathbf{V} = [\mathbf{v}_1, \cdots, \mathbf{v}_R] \in \mathbb{R}^{d \times R}$ for $R$ region proposals in the image.\footnote{The way of extracting word and region features is described in Sec~\ref{sec:implementdetail}.}
Our objective is to calculate the matching score between images and sentences based on $\mathbf{V}$ and $\mathbf{T}$. 
%


\subsection{Cross-modal Message Aggregation}
\label{subsec:attention_integration}

We propose a Cross-modal Message Aggregation module which aggregates the messages to be passed between regions and words. 
The aggregated message is obtained by a cross-modal attention mechanism, where the model takes the information from the other modality as cues to attend to the information from the self modality. 
In particular, our model performs word-level attention based on the cues from region features, and performs region-level attention based on the cues from word features. 
Such a message aggregation enables the information flow between textual and visual information, and the cross-modal attention for aggregating messages selects the most salient cross-modal information specifically for each word/region.
%
%

\setlength{\belowdisplayskip}{4pt}
\setlength{\belowdisplayshortskip}{4pt}
\setlength{\abovedisplayskip}{4pt}
\setlength{\abovedisplayshortskip}{4pt}

Mathematically, we first project region features and word features to a low dimensional space, and then compute the region-word affinity matrix,
\begin{equation}
\label{eq:AffinityMatrix} 
  \mathbf{A} = (\mathbf{\tilde{W}}_v\mathbf{V})^\top(\mathbf{\tilde{W}}_t\mathbf{T}), 
\end{equation}
where $\mathbf{\tilde{W}}_v, \mathbf{\tilde{W}}_s\in\mathbb{R}^{d_h \times d}$ are projection matrices which project the $d$-dimensional region or word features into a $d_h$-dimensional space. 
$\mathbf{A} \in \mathbb{R}^{R \times N}$ is the region-word affinity matrix where $\mathbf{A}_{ij}$ represents the affinity between the $i$th region and the $j$th word.
To derive the attention on each region with respect to each word, we normalize the affinity matrix over the image region dimension to obtain a word-specific region attention matrix,
\begin{equation}
\mathbf{\tilde{A}}_v = {\rm softmax}(\frac{\mathbf{A}^\top}{\sqrt{d_h}}),
\end{equation}
where the $i$th row of $\mathbf{\tilde{A}}_v$ is the attention over all regions with respect to the $i$th word.
We then aggregate all region features with respect to each word based on the word-specific region attention matrix,
\begin{align}
 \tilde{\mathbf{V}} &= \mathbf{\tilde{A}}_v\mathbf{V}^\top,
\end{align}
where the $i$th row of $\mathbf{\tilde{V}} \in \mathbb{R}^{N \times d}$ denotes the visual features attended by the $i$th word.

Similarly, we can calculate the attention weights on each word with respect to each image region, by normalizing the affinity matrix $\mathbf{A}$ over the word dimension. 
And based on the region-specific word attention matrix $\mathbf{\tilde{A}}_s$, we aggregate the word features to obtain the textual features attended by each region $\mathbf{\tilde{T}} \in \mathbb{R}^{R \times d}$,
\begin{align}
\vspace{-5pt}
\mathbf{\tilde{A}}_t = {\rm softmax}(\frac{\mathbf{A}}{\sqrt{d_h}}),~~~\tilde{\mathbf{T}} &= \mathbf{\tilde{A}}_t\mathbf{T}^\top.
\vspace{-5pt}
\end{align}

Intuitively, the $i$th row of $\mathbf{\tilde{V}}$ represents the visual features corresponding to the $i$th word, and the $j$th row of $\mathbf{\tilde{T}}$ represents the textual features corresponding to the $j$th region. Such a message aggregation scheme takes cross-modal interactions into consideration. $\tilde{\mathbf{V}}$ and $\tilde{\mathbf{T}}$ are the aggregated messages to be passed from visual features to textual features, and from textual features to visual features, respectively.

%


\subsection{Cross-modal Gated Fusion}
\label{subsec:gate_fusion}
The Cross-modal Message Aggregation module aggregates the most salient cross-modal information for each word/region as messages to be passed between textual and visual modalities, and the process of aggregating messages enables the interactions between modalities. 
However, with such a mechanism, the word and region features are still aggregated from each modality separately, without being fused together. 
To explore deeper and more complex interactions between images and sentences, the next challenge we face is how to fuse the information from one modality with the messages passed from the other modality. 

\begin{figure}[t]
\begin{center}
   \includegraphics[width=1\linewidth]{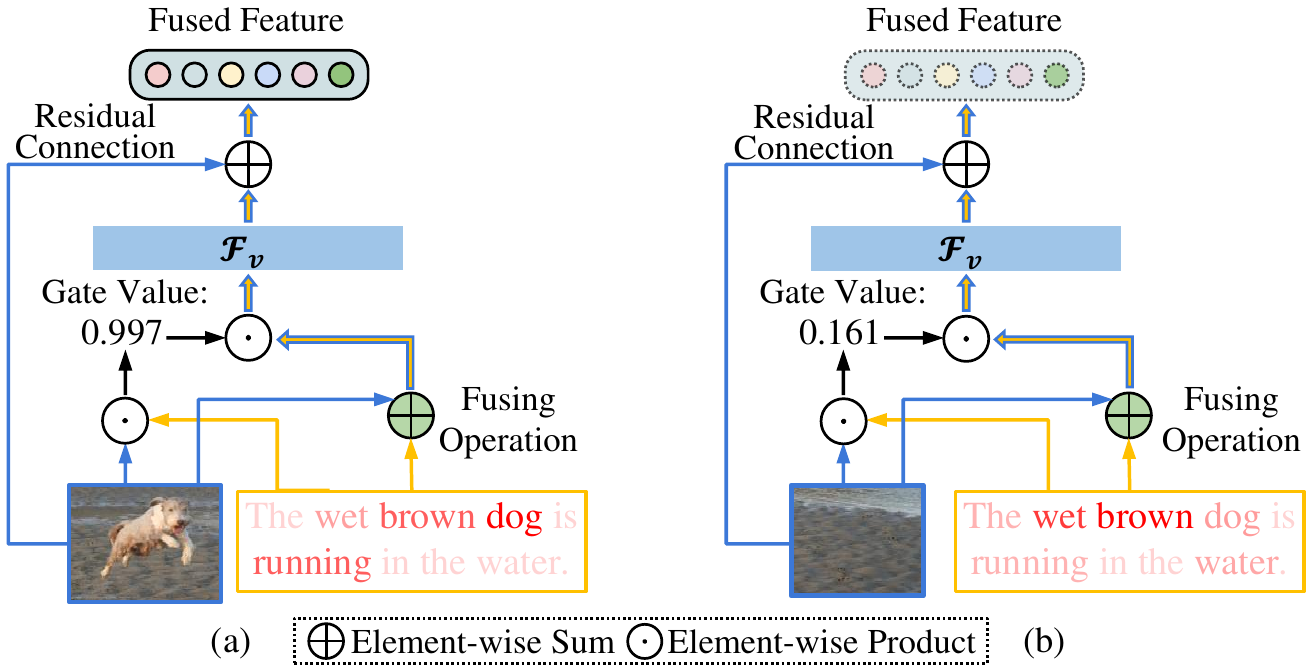}
\end{center}
\vspace{-0.2cm}
   \caption{Illustration of the fusion between original region features and aggregated textual messages for the Cross-modal Gated Fusion module. (a) denotes the fusion of a positive region and textual message pair, and (b) denotes the fusion of a negative region and textual message pair.}
\label{fig:layers}
\vspace{-5mm}
\end{figure}

However, conventional fusion operation assumes that the visual and textual features are matched, which is not the case for text-image retrieval. Directly fusing between the negative (mismatched) image-sentence pairs may lead to meaningless fused representation and may impede training and inference. 
Experiments also indicate that fusing the negative image-sentence pairs degrades the performance.
To this end, we design a novel \textit{Cross-modal Gated Fusion} module, as shown in Fig.~\ref{fig:layers}, to adaptively control the cross-modal feature fusion. 
More specifically, we want to fuse textual and visual features to a large extent for matched pairs, and suppress the fusion for mismatched pairs.
%

By the aforementioned Cross-modal Adaptive Message Passing module, we obtain the aggregated message $\mathbf{\tilde{V}}$ passed from visual to textual modality, and the aggregated message $\mathbf{\tilde{T}}$ passed from textual to visual modality. 
Our Cross-modal Gated Fusion module fuses $\mathbf{\tilde{T}}$ with the original region-level visual features $\mathbf{V}$ and fuses $\mathbf{\tilde{T}}$ with the original word-level textual features $\mathbf{T}$. 
%
%
%
We denote the fusion operation as $\oplus$ (\eg~element-wise add, concatenation, element-wise product). 
In practice, we use element-wise add as the fusion operation.
In order to filter out the mismatched information for fusion, a region-word level gate adaptively controls to what extent the information is fused. 

Take the fusion of original region features $\mathbf{V}$ and messages passed from the textual modality $\mathbf{\tilde{T}}$ as an example. 
Denote the $i$th region features as $\mathbf{v}_i$ (the $i$th column of $\mathbf{V}$), and denote the attended sentence features with respect to the $i$th region as $\mathbf{\tilde{t}}_i^\top$ (the $i$th row of $\mathbf{\tilde{T}}$). $\mathbf{\tilde{t}}_i^\top$ is the message to be passed from the textual modality to the visual modality. We calculate the corresponding gate as,
{\setlength\abovedisplayskip{8pt}
\setlength\belowdisplayskip{8pt}
\begin{align}
 \mathbf{g}_{i} = \sigma({\mathbf{v}_i \odot \mathbf{\tilde{t}}_i^\top }), \ i \in \{1, \cdots, R\}.
\end{align}
}
where $\odot$ denotes the element-wise product, $\sigma(\cdot)$ denotes the sigmoid function, and $ \mathbf{g}_{i} \in \mathbb{R}^{d}$ is the gate for fusing $\mathbf{v}_i$ and $\mathbf{\tilde{t}}_i^\top$. 
With such a gating function, if a region matches well with the sentence, it will receive high gate values which encourage the fusion operation. 
On the contrary, if a region does not match well with the sentence, it will receive low gate values, suppressing the fusion operation.
We represent the region-level gates for all regions as $\mathbf{G}_v = [\mathbf{g}_{1}, \cdots, \mathbf{g}_R] \in \mathbb{R}^{d \times R}$
%
We then use these gates to control how much information should be passed for cross-modality fusion. 
In order to preserve original information for samples that should not be intensively fused, the fused features are further integrated with the original features via a residual connection.
{{\setlength\abovedisplayskip{8pt}
\setlength\belowdisplayskip{8pt}
\begin{align}
 \mathbf{\hat{V}} &= \mathcal{F}_v(\mathbf{G}_v \odot (\mathbf{V}\oplus \mathbf{\tilde{T}}^\top)) + \mathbf{V},
\end{align}
}}
where $\mathcal{F}_v$ is a learnable transformation composed of a linear layer and non-linear activation function. $\odot$ denotes element-wise product, $\oplus$ is the fusing operation (element-wise sum), and $\mathbf{\hat{V}}$ is the fused region features. For positive pairs where the regions match well with the sentence, high gate values are assigned, and deeper fusion is encouraged. On the other hand, for negative pairs with low gate values, the fused information is suppressed by the gates, and thus $\mathbf{\hat{V}}$ is encouraged to keep the original features $\mathbf{V}$. Symmetrically, $\mathbf{T}$ and $\mathbf{\tilde{V}}$ can be fused to obtain $\mathbf{\hat T}$.

{\setlength\abovedisplayskip{0pt}
\setlength\belowdisplayskip{0pt}
\begin{align}
 \mathbf{h}_{i} &= \sigma({\mathbf{\tilde{v}}_i^\top \odot \mathbf{t}_i }), \ i \in \{1, \cdots, N\}, \\
 \mathbf{H}_t &= [\mathbf{h}_1, \cdots, \mathbf{h}_N] \in \mathbb{R}^{d \times N}, \\
 \mathbf{\hat{T}} &= \mathcal{F}_t(\mathbf{H}_t \odot (\mathbf{T}\oplus \mathbf{\tilde{V}}^\top)) + \mathbf{T}.
\end{align}
}


\vspace{-3mm}
\subsection{Fused Feature Aggregation for Cross-modal Matching} 
\label{subsec:fused_agg}
%
%


We use a simple attention approach to aggregate the fused features of $R$ regions and $N$ words into feature vectors representing the whole image and the whole sentence.
Specifically, given the fused features $\mathbf{\hat{V}} \in \mathbb{R}^{d \times R}$ and $\mathbf{\hat{T}} \in \mathbb{R}^{d \times N}$, the attention weight matrix is calculated by a linear projection and SoftMax normalization, and we aggregate the region features with the attention weights.
\begin{align}
 \mathbf{a}_{v} &= {\rm softmax}\big(\frac{\mathbf{W}_{v} \mathbf{\hat V}}{\sqrt{d}}\big)^\top,&\mathbf{v^*} &= \mathbf{\hat V}\mathbf{a}_{v}. \\
 \mathbf{a}_{t} &= {\rm softmax}\big(\frac{\mathbf{W}_{t} \mathbf{\hat T}}{\sqrt{d}}\big)^\top,&\mathbf{t^*} &= \mathbf{\hat T}\mathbf{a}_{t}.
\end{align}
where $\mathbf{W}_{v}, \mathbf{W}_{t} \in \mathbb{R}^{1 \times d}$ denotes the linear projection parameters, and $\mathbf{a}_{v} \in \mathbb{R}^{R}$ denotes the attention weights for the fused feature of $R$ regions, and $\mathbf{a}_{t} \in \mathbb{R}^{N}$ denotes the attention weights for the fused feature of $N$ words.
$\mathbf{v}^* \in \mathbb{R}^{d}$ is the aggregated features representation from $\mathbf{\hat{V}}$, and $\mathbf{t}^* \in \mathbb{R}^{d}$ is the aggregated features representation from $\mathbf{\hat{R}}$.

%

\subsection{Infer Text-image Matching with Fused Features}
\label{sec:infer}

Most previous approaches for text-image matching exploit visual-semantic embedding, which map the images and sentences into a common embedding space and calculates their similarities in the joint space~\cite{lee2018stacked, faghri2017vse++,gu2018look,zheng2017dual,nam2016dual}. 
Generally, consider the sampled positive image-sentence pair $(\mathcal{I},\mathcal{C})$ and negative image-sentence pairs $(\mathcal{I}, \mathcal{C^{'}})$, $(\mathcal{I^{'}}, \mathcal{C})$, the visual-semantic alignment is manipulated by the ranking loss with hardest negatives,
\begin{multline}
\label{eq:rankingloss}
\mathcal{L}_{rank-h}(\mathcal{I},\mathcal{C}) = \max_{\mathcal{C^{'}}}[\alpha-m(\mathcal{I},\mathcal{C})+m(\mathcal{I},\mathcal{C^{'}})]_+ \\
 + \max_{\mathcal{I^{'}}}[\alpha-m(\mathcal{I},\mathcal{C})+m(\mathcal{I^{'}},\mathcal{C})]_+,
\end{multline}
where $m(\mathcal{I},\mathcal{C})$ denotes the matching score, which is calculated by the distance of features in the common embedding space. 
$[x]_+=\max(0,x)$, $\alpha$ is the margin for ranking loss, and $\mathcal{C^{'}}$ and $\mathcal{I^{'}}$ are negative sentences and images, respectively.

With our proposed cross-modal Cross-modal Adaptive Message Passing model, however, the fused features can no longer be regarded as separate features in the same embedding space. 
Thus we cannot follow conventional visual-semantic embedding assumption to calculate the cross-modal similarities by feature distance in the joint embedding space. 
Instead, given the aggregated fused features $v^*$ and $s^*$, we re-formulate the text-image matching as a classification problem (\ie~``match'' or ``mismatch'') and propose a hardest negative cross-entropy loss for training.
Specifically, we use a two-layer MLP followed by a sigmoid activation to calculate the final matching scores between images and sentences,
\begin{align}
m(\mathcal{I},\mathcal{C}) = \sigma({\rm{MLP}}(\mathbf{v^*}+\mathbf{t^*})).
\end{align}

Although ranking loss has been proven effective for joint embedding, it does not perform well for our fused features. 
We exploit a hardest negative binary cross-entropy loss for training supervision. 
%
\begin{multline}
\label{eq:bce_loss}
   \mathcal{L}_{BCE-h}(\mathcal{I}, \mathcal{C}) = \underbrace{\log(m(\mathcal{I},\mathcal{C})) + \max_{\mathcal{C^{'}}}[\log(1- m(\mathcal{I},\mathcal{C^{'}}))]}_\text{image-to-text matching loss} \\ 
    + \underbrace{\log(m(\mathcal{I},\mathcal{C})) + \max_{\mathcal{I^{'}}}[\log(1- m(\mathcal{I^{'}},\mathcal{C}))]}_\text{text-to-image matching loss},
\end{multline}
where the first term is the image-to-text matching loss, and the second term is the text-to-image matching loss. We only calculate the loss of positive pairs and the hardest negative pairs in a mini-batch. Experiments in ablation study in Sec.~\ref{sec:ablation} demonstrates the effectiveness of this loss.

%
In fact, projecting the comprehensive features from different modalities into the same embedding space is difficult for cross-modal embedding, and the complex interactions between different modalities cannot be easily described by a simple embedding. 
However, our problem formulation based on the fused features do not require the image and language features to be embedded in the same space, and thus encourages the model to capture more comprehensive and fine-grained interactions from images and sentences.

\section{Experiments}

\label{sec:experiment}
	

\subsection{Implementation Details}
\label{sec:implementdetail}

\noindent\textbf{Word and region features.}
We describe how to extract the region-level visual features $\mathbf{V} = [\mathbf{v}_1, \cdots, \mathbf{v}_R]$ and word-level sentence features $\mathbf{T} = [\mathbf{t}_1, \cdots, \mathbf{t}_N]$.

We exploit the Faster R-CNN~\cite{ren2015faster} with ResNet-101 to pretrained by Anderson~\etal~\cite{anderson2017bottom} to extract the top 36 region proposals for each image. A feature vector $\mathbf{m}_i \in \mathbb{R}^{2048}$ for each region proposal is calculated by average-pooling the spatial feature map. We obtain the $1024$-dimentional region features with a linear projection layer,
\vspace{-2pt}
\begin{align}
   \mathbf{v}_i = \mathbf{W_I}\mathbf{m}_i+\mathbf{b_I},
\end{align}
\vspace{-2pt}
where $\mathbf{W_I}$ and $\mathbf{b_I}$ are model parameters, and $\mathbf{v}_i$ is the visual feature for the $i$th region.

Given an input sentence with $N$ words, we first embed each word to a 300-dimensional vector $x_i, i\in \{1,\cdots,N\}$ and then use a single-layer bidirectional GRU~\cite{chung2014GRU} with $1024$-dimensional hidden states to process the whole sentence,
\vspace{-2pt}
\begin{align}\ \  
	\overrightarrow{\mathbf{h}_i} =  \overrightarrow{{\rm GRU}}(\overrightarrow{\mathbf{h}_{i-1}}, \mathbf{x}_i)&, \  
	\overleftarrow{\mathbf{h}_i} =  \overleftarrow{{\rm GRU}}(\overleftarrow{\mathbf{h}_{i+1}}, \mathbf{x}_i).
\end{align}
\vspace{-2pt}
The feature of each word is represented as the average of hidden states from the forward GRU and backward GRU,
\vspace{-2pt}
\begin{align}
   \mathbf{t}_i = \frac{\overrightarrow{\mathbf{h}_i}+\overleftarrow{\mathbf{h}_i}}{2}, i\in \{1,\cdots,N\}
\end{align}
%
In practice, we set the maximum number of words in a sentences as $50$. We clip the sentences which longer than the maximum length, and pad sentences with less than $50$ words with a special padding token.

\noindent\textbf{Training strategy.}
%
%
%
%
%
Adam optimizer is adopted for training. The learning rate is set to 0.0002 for the first 15 epochs and 0.00002 for the next 25 epochs. Early stopping based on the 
validation performance is used to choose the best model.

\subsection{Experimental Settings}
\label{sec:expsetting}

\noindent\textbf{Datasets.}
We evaluate our approaches on two widely used text-image retrieval datasets, Flickr30K~\cite{young2014image} and COCO~\cite{lin2014microsoft}.  
Flickr30K dataset contains 31,783 images where each image has 5 unique corresponding sentences. Following~\cite{karpathy2015deep,faghri2017vse++}, we use 1,000 images for validation and 1,000 images for testing.
COCO dataset contains 123,287 images, each with 5 annotated sentences. The widely used Karpathy split~\cite{karpathy2015deep} contains 113,287 images for training, 5000 images for validation and 5000 images for testing. Following the most commonly used evaluation setting, we evaluate our model on both the 5 folds of 1K test images and the full 5K test images.

\noindent\textbf{Evaluation Metrics.}
For text-image retrieval, the most commonly used evaluation metric is {\rm R@K}, which is the abbreviation for recall at $K$ and is defined as the proportion of correct matchings in top-k retrieved results.
We adopt {\rm R@1, R@5} and {\rm R@10} as our evaluation metrics.

\subsection{Quantitative Results}
\label{sec:quantresult}

\begin{table}
\begin{center}
\small
\setlength{\tabcolsep}{1.3mm}
\begin{tabular}{|l|ccc|ccc|}
\hline
\multicolumn{7}{|c|}{COCO 1K test images}\\
\hline
& \multicolumn{3}{c|}{Caption Retrieval}
& \multicolumn{3}{c|}{Image Retrieval} \\
{Method}
& {R@1} & {R@5} & {R@10}
& {R@1} & {R@5} & {R@10}  \\
\hline\hline
Order~\cite{vendrov2015order} & 46.7 & - & 88.9 & 37.9 & - & 85.9 \\
DPC~\cite{zheng2017dual} & 65.6 & 89.8 & 95.5 & 47.1 & 79.9 & 90.0 \\
VSE++~\cite{faghri2017vse++} & 64.6 & - & 95.7 & 52.0 & - & 92.0  \\
GXN~\cite{gu2018look} & 68.5 & - & 97.9 & 56.6 & - & 94.5 \\
SCO~\cite{huang2017learning} & 69.9 & 92.9 & 97.5 & 56.7 & 87.5 & 94.8 \\
CMPM~\cite{ying2018CMPM} & 56.1 & 86.3 & 92.9 & 44.6 & 78.8 & 89.0 \\
SCAN t-i~\cite{lee2018stacked} & 67.5 & 92.9 & 97.6 & 53.0 & 85.4 & 92.9 \\
SCAN i-t~\cite{lee2018stacked} & 69.2 & 93.2 & 97.5 & 54.4 & 86.0 & 93.6 \\
\hline
CAMP (ours) & \textbf{72.3} & \textbf{94.8} & \textbf{98.3} & \textbf{58.5} & \textbf{87.9} & \textbf{95.0} \\
\hline
\multicolumn{7}{c}{} \\
\hline
\multicolumn{7}{|c|}{COCO 5K test images}\\
\hline
& \multicolumn{3}{c|}{Caption Retrieval}
& \multicolumn{3}{c|}{Image Retrieval} \\
{Method}
& {R@1} & {R@5} & {R@10}
& {R@1} & {R@5} & {R@10}  \\
\hline\hline
Order~\cite{vendrov2015order} & 23.3 & - & 84.7 & 31.7 & - & 74.6 \\
DPC~\cite{zheng2017dual} & 41.2 & 70.5 & 81.1 & 25.3 & 53.4 & 66.4 \\
VSE++~\cite{faghri2017vse++} & 41.3 & - & 81.2 & 30.3 & - & 72.4  \\
GXN~\cite{gu2018look} & 42.0 & - & 84.7 & 31.7 & - & 74.6 \\
SCO~\cite{huang2017learning} & 42.8 & 72.3 & 83.0 & 33.1 & 62.9 & 75.5 \\
CMPM~\cite{ying2018CMPM} & 31.1 & 60.7 & 73.9 & 22.9 & 50.2 & 63.8 \\
SCAN i-t~\cite{lee2018stacked}& 46.4 & 77.4 & 87.2 & 34.4 & 63.7 & 75.7 \\
\hline
CAMP (ours) & \textbf{50.1} & \textbf{82.1} & \textbf{89.7} & \textbf{39.0} & \textbf{68.9} & \textbf{80.2} \\
\hline
\end{tabular}
\end{center}
\vspace{-1.5mm}
\caption{Results by CAMP and compared methods on COCO.}
\label{tab:coco}
\end{table}

\begin{figure*}
\vspace{-10pt}
\begin{center}
   \includegraphics[width=1.0\linewidth]{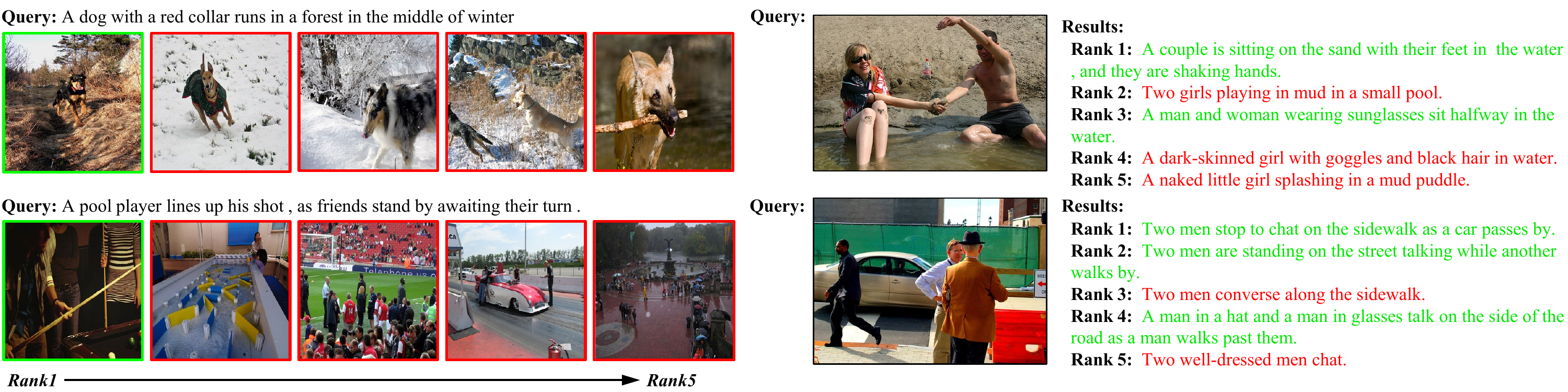}
\end{center}
\vspace{-2.5mm}
   \caption{Qualitative retrieval results. The top-5 retrieved results are shown. Green denotes the ground-truth images or captions. Our model is able to capture the comprehensive and fine-grained alignments between images and captions by incorporating cross-modal interactions.}
\label{fig:results}
\vspace{-5mm}
\end{figure*}

\begin{table}
\begin{center}
\small
\setlength{\tabcolsep}{1.3mm}
\begin{tabular}{|l|ccc|ccc|}
\hline
\multicolumn{7}{|c|}{Flickr30K 1K test images}\\
\hline
& \multicolumn{3}{c|}{Caption Retrieval}
& \multicolumn{3}{c|}{Image Retrieval} \\
{Method}
& {R@1} & {R@5} & {R@10}
& {R@1} & {R@5} & {R@10}  \\
\hline\hline
VSE++~\cite{faghri2017vse++} & 52.9 & - & 87.2 & 39.6 & - & 79.5 \\
DAN~\cite{nam2016dual} & 55.0 & 81.8 & 89.0 & 39.4 & 69.2 & 79.1 \\
DPC~\cite{zheng2017dual} & 55.6 & 81.9 & 89.5 & 39.1 & 69.2 & 80.9 \\
SCO~\cite{huang2017learning} & 55.5 & 82.0 & 89.3 & 41.1 & 70.5 & 80.1 \\
CMPM~\cite{ying2018CMPM} & 49.6 & 76.8 & 86.1 & 37.3 & 65.7 & 75.5 \\
SCAN t-i~\cite{lee2018stacked} & 61.8 & 87.5 & 93.7 & 45.8 & 74.4 & 83.0 \\
SCAN i-t~\cite{lee2018stacked} & 67.7 & 88.9 & 94.0 & 44.0 & 74.2 & 82.6 \\
\hline
CAMP (ours) & \textbf{68.1} & \textbf{89.7} & \textbf{95.2} & \textbf{51.5} & \textbf{77.1} & \textbf{85.3} \\
\hline
\end{tabular}
\end{center}
\vspace{-1.5mm}
\caption{Results by CAMP and compared methods on Flickr30K.}
\label{tab:f30k}
\vspace{-5mm}
\end{table}

Table~\ref{tab:coco} presents our results compared with previous methods on 5k test images and 5 folds of 1k test images of COCO dataset, respectively. Table~\ref{tab:f30k} shows the quantitative results on Flickr30k dataset of our approaches and previous methods. VSE++~\cite{faghri2017vse++} jointly embeds image features and sentence features into the same embedding space and calculates image-sentence similarities as distances of embedded features, and train the model with ranking loss with hardest negative samples in a mini-batch. SCAN~\cite{lee2018stacked} exploits stacked cross attention on either region features or word features, but does not consider message passing or fusion between image regions and words in sentences. Note that the best results of SCAN~\cite{lee2018stacked} employ an ensemble of two models. For fair comparisons, we only report their single model results on the two datasets. 

Experimental results show that our Cross-modal Adaptive Message Passing (CAMP) model outperforms previous approaches by large margins, demonstrating the effectiveness and necessity of exploring the interactions between visual and textual modalities for text-image retrieval. 






\vspace{-1mm}
\subsection{Qualitative Results}
\label{sec:qualitative}

\begin{figure*}[t]
\begin{center}
   \includegraphics[width=0.9\linewidth]{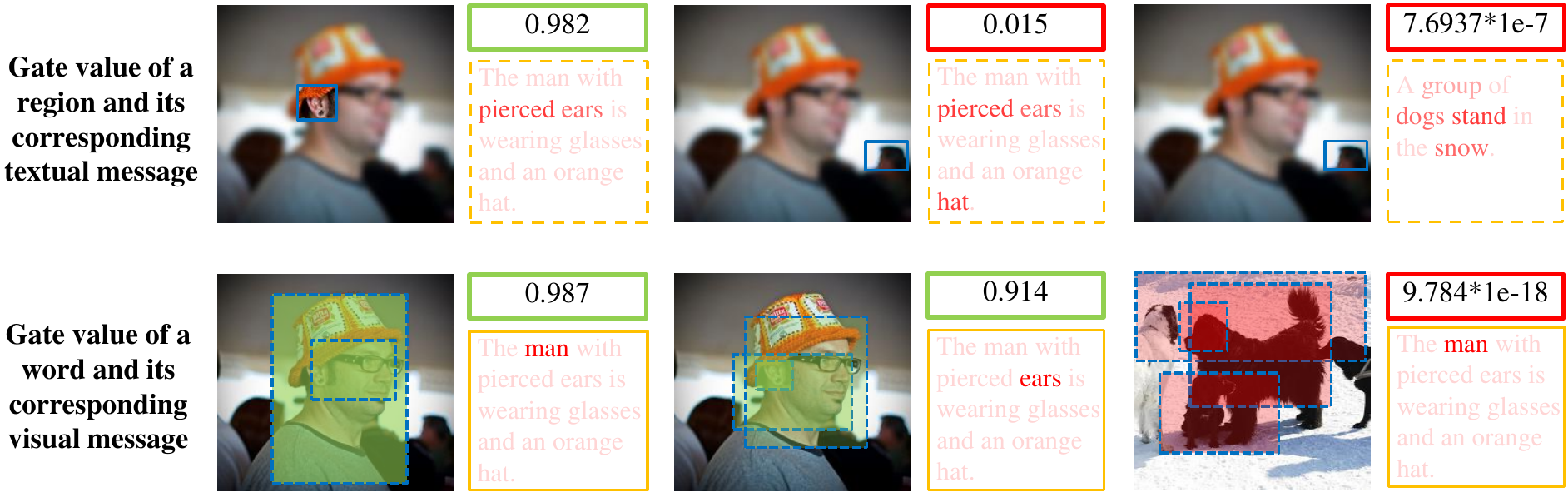}
\end{center}
\vspace{-10pt}
   \caption{Gate values for aggregated textual/visual messages and original regions/words. High gate values indicate strong textual-visual alignments, encouraging deep cross-modal fusion. Low gate values suppress the fusion of uninformative regions or words for matching.}
\label{fig:visgate}
\vspace{-5mm}
\end{figure*}

\vspace{-0.5mm}
We show qualitative results by our gated fusion model for text-to-image and image-to-text retrieval in Fig.~\ref{fig:results}. 
Take images in the first row of the left part as an example.
We retrieve images based on the query caption ``A dog with a red collar runs in a forest in the middle of winter.'' 
Our model successfully retrieves the ground-truth image. 
Note that the all of the top 5 retrieved images all related to the query caption, but the top 1 image matches better in details such as ``runs in a forest'' and ``red collar''.
By alternatively attending to, passing messages and fusing between both modalities to incorporate deep cross-modal interactions, the model would have the potential of discovering such fine-grained alignments between images and captions.
%



\subsection{Ablation Study}
\label{sec:ablation}
\setlength{\tabcolsep}{2pt}
\begin{table}[t]
\footnotesize
\begin{center}
\begin{tabular}{|l|ccc|ccc|}
\hline
 \multicolumn{7}{|c|}{Ablation study results on Flickr30K} \\
\hline
& \multicolumn{3}{c|}{Caption Retrieval}
& \multicolumn{3}{c|}{Image Retrieval} \\

{Method}
& {R@1} & {R@5} & {R@10}
& {R@1} & {R@5} & {R@10}  \\
\hline\hline
CAMP & \textbf{68.1} & \textbf{89.7} & \textbf{95.2} & \textbf{51.5} & \textbf{77.1} & \textbf{85.3} \\
\hline
Base model & 63.5 & 87.1 & 93.1 & 46.2 & 74.2 & 83.4 \\
w/o cross-attn & 59.7 & 83.5 &  88.9 & 41.2 & 65.5 & 79.1 \\
w/o fusion & 65.6 & 88.0 & 94.9 & 48.2 & 75.7 & 84.9 \\
Fusion w/o gates & 61.7 & 86.3 & 92.6 & 45.1 & 72.1 & 80.7 \\
Fusion w/o residual & 56.7 & 83.9 & 91.5 & 43.7 & 72.6 & 79.3 \\
w/o attn-based agg & 63.4 & 86.8 & 93.5 & 47.5 & 73.1 & 82.8 \\
Concat fsuion & 66.3 & 89.0 & 94.3 & 51.0 & 74.1 & 83.3 \\
Product fusion & 61.5 & 87.3 & 93.2 & 49.9 & 74.0& 80.5 \\
Joint embedding & 62.0 & 87.8 & 92.4 & 46.3 & 73.7 & 80.3 \\
MLP+Ranking loss & 60.9 & 87.5 & 92.4 & 44.3 & 70.1 & 79.4 \\
BCE w/o hardest & 65.5 & 89.1 & 94.6 & 50.8 & 76.1 & 83.2 \\
\hline

\end{tabular}
\end{center}
\vspace{-2mm}
\caption{Results of ablation studies on Flickr30K .}
\label{tab:ablation}
\vspace{-5mm}
\end{table}

Our carefully designed Cross-modal Adaptive Message Passing model has shown superior performance, compared with conventional approaches that independently embed images and sentences to the joint embedding space without fusion. We carry several ablation experiments to validate the effectiveness of our design.


\noindent\textbf{Base model without Cross-modal Adaptive Message Passing.}
To illustrate the effectiveness of our model, we design a baseline model without any cross-modal interactions.
The baseline model attends to region features and word features separately to extract visual and textual features, and compare their similarities by cosine distance. The detailed structure is provided in the supplementary material.
Ranking loss with hardest negatives is used as training supervision.
The results are shown as ``Base model'' in Table~\ref{tab:ablation}, indicating that our CAMP model improves the base model without interaction by a large margin.

\noindent\textbf{The effectiveness of cross-modal attention for Cross-modal Message Aggregation.}
In the Cross-modal Message Aggregation module, we aggregate messages to be passed to the other modality by cross-modal attention between two modalities. 
We experiment on removing the cross-modal attention and simply average the region or word features, and using the average word/region features as aggregated messages.
Results are shown as ``w/o cross-attn'' in Table~\ref{tab:ablation}, indicating that removing the cross-modal attention for message aggregation would decrease the performance.
We visualize some examples of cross-modal attention in the supplementary material.

\noindent\textbf{The effectiveness of Cross-modal Gated Fusion.}
We implement a cross-modal attention model without fusion between modalities. 
The cross-modal attention follows the same way as we aggregate cross-modal messages for message passing in Sec.~\ref{subsec:attention_integration}.
Text-to-image attention and image-to-text attention are incorporated symmetrically.
It has the potential to incorporate cross-modal interactions by attending to a modality with the cue from another modality, but no cross-modal fusion is adopted.
The detailed structures are provided in the supplementary material.
By comparing the performance of this model (denoted as ``w/o fusion'' in Table~\ref{tab:ablation}) with our CAMP model, we demonstrate that cross-modal fusion is effective in incorporating deeper cross-modal interactions.
Additionally, the average gate values for positive and negative pairs are $0.971$ and $2.7087*10^{-9}$, respectively, indicating that the adaptive gates are able to filter out the mismatched information and encourage fusion between aligned information.

\noindent\textbf{The necessity of adaptive gating and residual connection for Cross-modal Gated Fusion.}
We propose the adaptive gates to control to what extent the cross-modality information should be fused. 
Well-aligned features are intensively fused, while non-corresponding pairs are slightly fused. 
Moreover, there is a residual connection to encourage the model to preserve the original information if the gate values are low.
We conduct experiments on fusion without adaptive gates or residual connection, denoted by ``Fusion w/o gates'' and ``Fusion w/o residual'' in Table~\ref{tab:ablation}. Also, to show the effectiveness of our choice among several fusion operations, two experiments denoted as ``Concat fusion'' and ``Product fusion'' are conducted to show the element-wise addition is slightly better.
Results 
indicate that using a conventional fusion would confuse the model and cause a significant decline in performance.
Moreover, we show some examples of gate values in Fig.~\ref{fig:visgate}. Words/regions that are strongly aligned to the image/sentence obtains high gate values, encouraging the fusing operation. While the low gate values would suppress the fusion of uninformative regions or words for matching.
%
%
Note that the gate values between irrelevant background information may also be low even though the image matches with the sentence.
In this way, the information from the irrelevant background is suppressed, and the informative regions are highlighted.

\noindent\textbf{The effectiveness of attention-based fused feature aggregation.}
In Sec.~\ref{subsec:fused_agg}, a simple multi-branch attention is adapted to aggregate the fused region/word-level features into a feature vector representing the whole image/sentence. 
We replace this attention-based fused feature aggregation with a simple average pooling along region/word dimension. 
Results denoted as ``w/o attn-based agg'' show the effectiveness of our attention-based fused feature aggregation.

\noindent\textbf{Different choices for inferring text-image matching score and loss functions.}
Since the fused features cannot be regarded as image and sentence features embedded in the joint embedding space anymore, they should not be matched by feature distances. 
In Sec.~\ref{sec:infer}, we reformulate the matching problem based on the fused features, by predicting the matching score with MLP on the fused features, and adopting hardest negative cross-entropy loss as training supervision.
In the experiment denoted as ``joint embedding'' in Table~\ref{tab:ablation}, we follow conventional joint embedding approaches to calculate the matching score by cosine distance of the fused features $\hat{s}$ and $\hat{v}$, and employ the ranking loss (Eq.\eqref{eq:rankingloss}) as training supervision.
In the experiment denoted as ``MLP+ranking loss'', we use MLP on the fused features to predict the matching score, and adopt ranking loss for training supervision.
We also test the effectiveness of introducing hardest negatives in a mini-batch for cross-entropy loss. 
In the experiment denoted as ``BCE w/o hardest'', we replace our hardest negative BCE loss with the conventional BCE loss without hardest negatives,
%
where $b$ is the number of negative pairs in a mini-batch, to balance the loss of positive pairs and negative pairs.
Those experiments show the effectiveness of our scheme for predicting the matching score based on the fused features, and validates our hardest negative binary cross-entropy loss designed for training text-image retrieval.

\vspace{-3mm}

\section{Conclusion} 
\vspace{-2mm}
Based on the observation that cross-modal interactions should be incorporated to benefit text-image retrieval, we introduce a novel Cross-modal Gated Fusion (CAMP) model to adaptively pass messages across textual and visual modalities. Our approach incorporates the comprehensive and fine-grained cross-modal interactions for text-image retrieval, and properly deals with negative (mismatched) pairs and irrelevant information with an adaptive gating scheme. We demonstrate the effectiveness of our approach by extensive experiments and analysis on benchmarks.
\vspace{1mm}

\noindent \textbf{Acknowledgements}
This work is supported in part by SenseTime Group Limited, in part by the General Research Fund through the Research Grants Council of Hong Kong under Grants CUHK14202217, CUHK14203118, CUHK14205615, CUHK14207814, CUHK14213616, CUHK14208417, CUHK14239816, in part by CUHK Direct Grant.

{\small
\bibliographystyle{ieee_fullname}
\bibliography{bib}
}

\end{document}